\documentclass[10pt,twocolumn,letterpaper]{article}

\usepackage{ijcb}
\usepackage{times}
\usepackage{epsfig}
\usepackage{graphicx}
\usepackage{amsmath}
\usepackage{amssymb}
\usepackage[colorlinks,breaklinks,pagebackref]{hyperref}
\usepackage{standalone}
\usepackage{todonotes}
\usepackage{subcaption}
\usepackage{multirow}
\usepackage{arydshln}
\usepackage[most]{tcolorbox}
\usepackage{booktabs}
\usepackage[font={small}]{caption}
\usepackage{bm}
\usepackage[toc,page]{appendix}

\makeatletter
\@namedef{ver@everyshi.sty}{}
\makeatother

\ijcbfinalcopy 

\begin{document}

\title{\ourmethod{}: Homomorphically Encrypted Fusion of Biometric Templates}
\author{
Luke Sperling$^{\dagger}$
\quad
Nalini Ratha$^{\ddagger}$
\quad
Arun Ross$^{\dagger}$
\quad
Vishnu Naresh Boddeti$^{\dagger}$
\\
\vspace{2mm}
{$^{\dagger}$Michigan State University \quad $^{\ddagger}$University at Buffalo}
}

\newcommand{\ourmethod}{HEFT}
\newcommand{\vishnu}[1]{{\textcolor{red}{Vishnu: #1}}}

\maketitle


\begin{abstract}
    This paper proposes a non-interactive end-to-end solution for secure fusion and matching of biometric templates using fully homomorphic encryption (FHE). Given a pair of encrypted feature vectors, we perform the following ciphertext operations, i) feature concatenation, ii) fusion and dimensionality reduction through a learned linear projection, iii) scale normalization to unit $\ell_2$-norm, and iv) match score computation. Our method, dubbed \ourmethod{} (Homomorphically Encrypted Fusion of biometric Templates), is custom-designed to overcome the unique constraint imposed by FHE, namely the lack of support for non-arithmetic operations. From an inference perspective, we systematically explore different data packing schemes for computationally efficient linear projection and introduce a polynomial approximation for scale normalization. From a training perspective, we introduce an FHE-aware algorithm for learning the linear projection matrix to mitigate errors induced by approximate normalization. Experimental evaluation for template fusion and matching of face and voice biometrics shows that \ourmethod{} (i) improves biometric verification performance by 11.07\% and 9.58\% AUROC compared to the respective unibiometric representations while compressing the feature vectors by a factor of 16 (512D to 32D), and (ii) fuses a pair of encrypted feature vectors and computes its match score against a gallery of size 1024 in 884 ms. Code and data are available at \href{https://github.com/human-analysis/encrypted-biometric-fusion}{https://github.com/human-analysis/encrypted-biometric-fusion}
\end{abstract}

\section{Introduction \label{sec:introduction}}
Feature-level fusion is a commonly employed technique in multi-biometric recognition systems, especially in large-scale deployments. Template fusion helps to overcome the limitations of unibiometric systems in terms of improving recognition performance and population coverage. However, utilizing multiple biometric signatures also enhances the security risks associated with attacks on such systems. In fact, there is growing evidence that the templates contain sufficient information to either reconstruct the raw biometric signature~\cite{mai2018reconstruction} or leak sensitive soft-biometric information~\cite{liu2015deep}. Thus, it is imperative to devise template fusion and matching schemes that secure the biometric signatures of users across all modalities and help protect their privacy. Realizing this goal is the primary focus of this paper.

\begin{figure*}[h]
    \centering
    \includegraphics[width=0.9\textwidth]{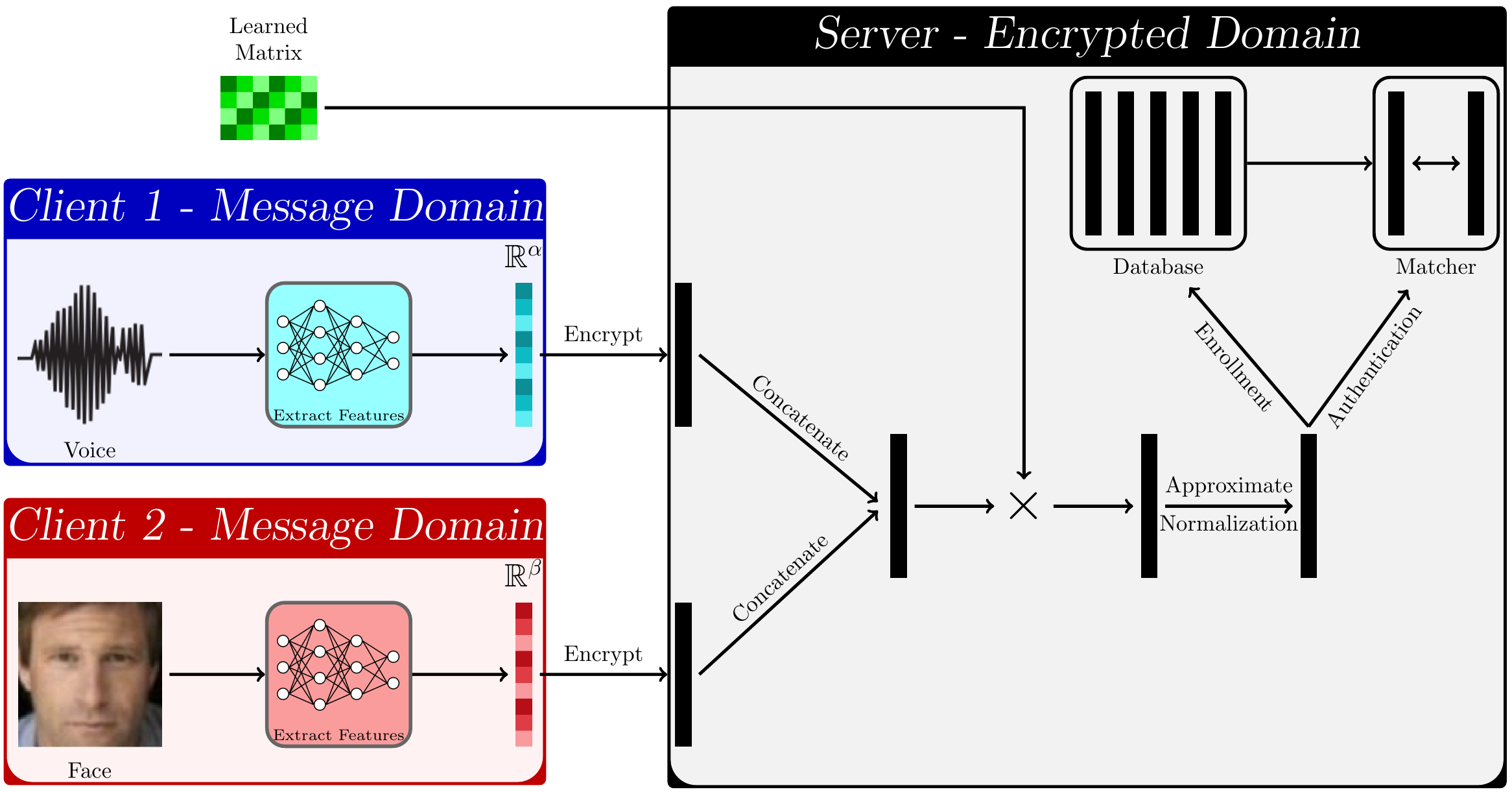}
    \caption{\textbf{Overview:} End-to-end biometric template fusion and matching using fully homomorphic encryption (FHE). Given feature representations extracted from two different modalities of an individual, the client encrypts and transmits the features to our system. We concatenate the two encrypted vectors and perform a matrix-vector multiplication with a learned plaintext projection matrix. The resulting ciphertext represents the fused encrypted vector. We normalize the encrypted vector using an approximation to overcome the constraints imposed by FHE. During enrollment, this template is stored in the database of encrypted templates. During authentication, match scores are computed between the probe and templates from the encrypted database and sent to the client for decryption and further processing.\label{fig:approach_overview}}
\end{figure*}

Cryptosystems based on Fully Homomorphic Encryption~\cite{gentry2012fully} (FHE) are an attractive solution for protecting biometric templates through encryption. FHE schemes such as BFV~\cite{brakerski2012fully, fan2012somewhat} and CKKS~\cite{cheon2017homomorphic}, theoretically, allow for computations directly on encrypted data without the need for decryption. Recent work~\cite{boddeti2018secure, engelsma2022hers} has demonstrated that FHE is exceptionally effective and scalable for securing biometric templates, allowing for encrypted matching and search against a gallery of 100 Million.

Template-level fusion and matching typically involve the following operations: feature concatenation, linear/non-linear projection, scale normalization of resulting feature, and finally matching score computation. Operations in existing approaches for feature-level fusion are all presumed to be performed in plaintext (unencrypted domain) and therefore run into limitations when performed on ciphertext (encrypted domain). For example, non-arithmetic operations, such as division and square root required for scale normalization, are not supported by FHE schemes for direct computation on ciphertexts. Furthermore, operations on ciphertext are significantly more computationally expensive, both in terms of latency and memory requirements, than the same operations on the corresponding plaintext. 

To overcome the aforementioned limitations, we propose \ourmethod{}, a biometric template fusion and matching scheme that operates directly on encrypted templates. Given a pair of encrypted templates, \ourmethod{} performs the following ciphertext operations: feature concatenation, projection, scale normalization to unit $\ell_2$-ball, and matching score computation. This process is illustrated in Fig. \ref{fig:approach_overview}.

From an \emph{inference perspective}, the salient features of \ourmethod{} include, i) fusing unibiometric templates of different dimensionalities, ii) fusion and compression of concatenated templates through linear projection to ease the steep computational burden of downstream ciphertext matching operations, and iii) approximating the non-arithmetic $\ell_2$ normalization operation through composite polynomials. From a \emph{learning perspective}, we introduce FHE-aware learning of fusion model parameters to mitigate performance loss from approximating the normalization process.

From a \emph{practical perspective}, we carefully analyzed the effect of various design choices on the trade-off between accuracy and efficiency (memory and latency) of biometric fusion and match score computation. These include data encoding schemes, matrix-multiplication methods, and approximation schemes for normalization. Through our analysis, we identify the optimal options (in terms of memory and latency) under both small-scale and large-scale settings, w.r.t. feature dimension and gallery size.

In summary, we present the first practically feasible homomorphic multibiometric feature-level fusion and matching algorithm. Experimental evaluation on a combination of encrypted face and voice biometric signatures demonstrates appreciable gains in matching performance over the unibiometric counterparts while taking 884 ms to fuse a pair of biometric templates and compute match scores against a gallery of size 1024.

\section{Related Work \label{sec:relatedworks}}
\noindent\textbf{Privacy-Preservation in Biometrics:} Many methods have been devised over the years to secure biometric templates and preserve user privacy. Early biometric cryptosystems based on image processing~\cite{soutar1998biometric, soutar1999biometric} and fuzzy vaults~\cite{juels2006fuzzy} were employed for protecting both iris~\cite{lee2008new} and fingerprint~\cite{uludag2005fuzzy} data. Such systems, however, suffered from a loss in matching performance. Cryptosystems such as Goldwasser-Micali encryption have also been used for authentication scenarios \cite{bringer2007application}, but they do not protect the templates at matching and are, therefore, vulnerable to attacks.

Homomorphic encryption (HE) is an attractive option for privacy-preserving biometrics applications due to its ability to enable computations on encrypted data without the need to decrypt. Early biometric systems driven by HE were based on partially homomorphic encryption (PHE) schemes~\cite{goldreich2009foundations}. They were applied to numerous biometric modalities~\cite{barni2015privacy}, including face recognition~\cite{troncoso2013fully}, iris recognition~\cite{upmanyu2009efficient,upmanyu2010blind, blanton2011secure} and fingerprint recognition~\cite{barni2010aprivacy}. The opportunity to design robust biometrics cryptosystems came to the fore with the development of the first fully homomorphic encryption (FHE) scheme~\cite{gentry2012fully}. Since then, there have been many application scenarios for biometrics exploiting the privacy afforded by FHE without substantial performance drawbacks.  Gomez-Barrero et al.~\cite{gomez2017multi} developed a general framework for template-level fusion based on homomorphic encryption. This framework relies on performing fusion before encryption and does not support template fusion directly in the encrypted domain. Boddeti~\cite{boddeti2018secure} demonstrated the ability to match face templates in the encrypted domain. Engelsma et al.~\cite{engelsma2022hers} proposed an efficient way to search encrypted templates by combining a novel encoding scheme with feature compression. By using a tree search structure created by fusing similar templates, Drozdowski et al.~\cite{drozdowski2021feature} developed a method for faster biometric indexing and retrieval. In contrast to this body of work, in this paper, we leverage fully homomorphic encryption for end-to-end template fusion and match score computation and devise an FHE-aware learning algorithm for feature projection.

\vspace{3pt}
\noindent\textbf{Feature-Level Biometric Fusion:} Fusion at the feature-level leverages information from multiple templates to improve performance. Early techniques focused on selecting features from each template to be fused~\cite{ross2005feature}. Sarangi et al.~\cite{sarangi2022feature} combined face and ear templates by concatenating templates compressed through classical dimensionality reduction techniques. Feature-level fusion has also been performed on face, fingerprint, and finger vein modalities~\cite{xin2018multimodal}. Coupled mapping techniques have been devised to match samples between domains, with a maximum-margin approach~\cite{siena2013maximum} and with a marginal fisher analysis approach~\cite{siena2012coupled}. Lately, learning-based approaches have been used. Silva et al.~\cite{silva2018multimodal} performed feature selection using Particle Swarm Optimization. Tiong et al.~\cite{tiong2020multimodal} proposed a method of information fusion via extracting features from raw biometric data using a CNN and then combining them with a series of fully connected layers.  Other deep learning approaches have been proposed recently \cite{bartuzi2018mobibits, zhang2018deep, alay2020deep, leghari2021deep, talreja2017multibiometric}. Contrasting these methods, we opt for a linear projection-based approach to limit the multiplicative depth of the circuit and decrease computational complexity, which is important for creating a practical solution in FHE.

\section{Approach\label{sec:approach}}
We propose \ourmethod{} for template fusion and matching. It is designed for maximizing performance and efficiency at inference over ciphertexts. Given an encrypted multibiometric dataset, i.e., a pair of encrypted feature vector matrices, \ourmethod{} performs the following series of ciphertext operations, (i) \emph{concatenation} of the feature vectors, (ii) \emph{linear projection} using a learned matrix to a new lower-dimensional feature space, (iii) \emph{approximate normalization} of the features by projecting them onto a unit $\ell_2$-ball for fast match score computation, and (iv) \emph{match score computation} of the fused features against an encrypted gallery of fused features. Finally, to compensate for the errors\footnote{leads to performance degradation in downstream tasks like matching.} induced by approximate normalization at inference, we propose an FHE-aware training process that takes the approximate normalization into account.

\subsection{Problem Setup}

\noindent\textbf{Biometric Fusion:} Consider a multibiometric system that comprises $n$ features vectors from two sources $\bm{X}=\left[\bm{x}_1,\dots,\bm{x}_n\right] \in \mathcal{R}^{\alpha \times n}$ and $\bm{Y}=\left[\bm{y}_1,\dots,\bm{y}_n\right] \in \mathcal{R}^{\beta \times n}$. These features are fused into a new space $\bm{Z}=\left[\bm{z}_1, \dots, \bm{z}_n\right] \in \mathcal{R}^{\gamma \times n}$. While we restrict ourselves to fusing a pair of biometric features, our solution can readily be applied to the fusion of a multitude of biometric features.

In this paper, we consider a linear projection operation to fuse the feature vectors, i.e., $\bm{Z} = \bm{P\tilde{X}}$, where $\bm{\tilde{X}} = \begin{bmatrix}\bm{x}_1 & \bm{x}_2 & \cdots & \bm{x}_n \\ \bm{y}_1 & \bm{y}_2 & \cdots & \bm{y}_n \end{bmatrix} \in \mathcal{R}^{\delta \times n}$ is a matrix of concatenated features and $\bm{P}\in\mathcal{R}^{\gamma \times \delta}$ is the projection matrix that maps into a common $\gamma$ dimensional space, and $\delta=\alpha + \beta$.

The fused templates can be used for any downstream tasks, such as matching. A common metric that is adopted for template matching is the cosine similarity $d(\bm{x},\bm{y}) = 1 - \frac{\bm{x}^T\bm{y}}{\|\bm{x}\|_2\|\bm{y}\|_2} = 1 - \bm{\tilde{x}}^T\bm{\tilde{y}}$ where $\bm{\tilde{x}}$ and $\bm{\tilde{y}}$ are scale normalized versions of $\bm{x}$ and $\bm{y}$, respectively, and are obtained by projecting $\bm{x}$ and $\bm{y}$ onto the unit $\ell_2$-ball.

\vspace{4pt}
\noindent\textbf{Secure Biometric Fusion:} Our goal in this work is to devise a cryptographic solution to secure the multibiometric templates and prevent unauthorized access to any private user information during the template fusion process, as well as any desired downstream tasks. This can be achieved through a parameterized function that transforms the multibiometric features $(\bm{x},\bm{y})$ into an alternate space $(\mathcal{E}(\bm{x}), \mathcal{E}(\bm{y}))$ such that $\mathcal{E}(\bm{x})=f(\bm{x};\bm{\theta}_{pk})$, $\bm{x}=g(\mathcal{E}(\bm{x});\bm{\theta}_{sk})$ are encryption and decryption functions with $\bm{\theta}_{pk}$ and $\bm{\theta}_{sk}$ being the public and secret keys respectively. By executing all the fusion operations, namely, \emph{concatenation}, \emph{projection}, \emph{normalization} and \emph{match score computation} directly over the ciphertexts, i.e., without decrypting them, we can prevent unauthorized access to sensitive information, and hence preserve user privacy.

Fully homomorphic encryption (FHE) is a class of encryption algorithms that allows arithmetic computations directly over ciphertexts and is ideally suited to realize our goal. Even if a malicious attacker can gain access to the multibiometric features during any part of the fusion or matching process, without access to the secret key $\bm{\theta}_{sk}$ the attacker cannot reconstruct the underlying biometric sample or extract any other information present in the features.

\subsection{Protocols: Template Fusion and Authentication}
We use the Cheon-Kim-Kim-Song (CKKS) scheme \cite{cheon2017homomorphic} as the underlying FHE scheme for template fusion and match score computation. We first give an overview of this scheme and describe the enrollment and authentication protocols for template fusion next.

The \textbf{CKKS encryption scheme} allows operations over encrypted vectors of complex numbers \cite{cheon2017homomorphic}. Its mathematical basis lies in modular arithmetic over polynomial rings, and its security lies in the hardness of the Ring Learning with Errors problem. CKKS offers post-quantum security for an appropriate choice of encryption parameters~\cite{albrecht2021homomorphic}. Plaintexts are polynomials within the polynomial ring $R = \mathbb{Z}[x]/(x^N + 1)$. Therefore, complex vectors $C^{N/2}$ must be encoded into this space to perform encryption. After encoding, the plaintext polynomial is encrypted via a secret key into a set of two polynomials, $R_q^2 = \mathbb{Z}_q[x]/(x^N + 1)$ where $R_q$ denotes polynomials of coefficients modulo $q$ and degree less than $N$. This will serve as the ciphertext.

CKKS has three keys, a secret key $sk$, a public key $pk$, and an evaluation key $evk$ for homomorphic multiplication. Its protocol comprises the following functions, i) \underline{\emph{Key Generation:}} Generates the keys, ii) \underline{\emph{Encryption:}} Given a plaintext polynomial and the public key, output two polynomials representing the ciphertext, iii) \underline{\emph{Decryption:}} Given a ciphertext comprised of two polynomials, apply the secret key and retrieve a plaintext polynomial, iv) \underline{\emph{Addition:}} A simple sum of the ciphertexts translates to homomorphic addition, v) \underline{\emph{Multiplication:}} Multiplication of ciphertexts is polynomial multiplication which results in three polynomials. To restrict the size of resultant ciphertexts, relinearization is needed, vi) \underline{\emph{Relinearization:}} Given three polynomials representing a ciphertext product, the evaluation key is used to reduce the size of the ciphertext from three to two polynomials, and vii) \underline{\emph{Rotation:}} Ciphertexts may be cyclically rotated using an optionally generated set of Galois keys.

\vspace{4pt}
\noindent\textbf{Encrypted Template Fusion Protocol at Enrollment:} Consider two sets of biometric templates $\bm{X}\in\mathcal{R}^{\alpha \times n}$ and $\bm{Y}\in\mathcal{R}^{\beta \times n}$ that we seek to fuse along with their class labels $\bm{I}\in\mathcal{Z}^{n}$. Each set of templates are encrypted using the data encoding scheme requested by the cloud server. After receiving the encrypted templates, the cloud server performs the following operations: i) for each class label $c$, create all pairs of templates $\{(\bm{x}_i, \bm{y}_j) | \forall (i,j) \in \bm{I}_c \times \bm{I}_c, \bm{I_c} \subseteq \bm{I} \}$, where $\bm{I}_c$ are the indices of samples belonging to class $c$, ii) fuse the pairs of templates created, i.e., \emph{concatenation}, \emph{projection} and \emph{normalization}, and iii) add the fused templates to the current gallery $\bm{G}$.

\vspace{4pt}
\noindent\textbf{Encrypted Template Fusion Protocol at Authentication:} A client sends a sample of encrypted multibiometric templates $\bm{x}\in\mathcal{R}^{\alpha}$ and $\bm{y}\in\mathcal{R}^{\beta}$. This pair of templates is fused, i.e., \emph{concatenation}, \emph{projection} and \emph{normalization} to create a probe template $\bm{z}\in\mathcal{R}^{\gamma}$. For identification, i.e., $1:N$ comparisons, match score (e.g., cosine similarity) is computed between the probe and the entire gallery $\bm{G}$. For verification with a claimed identity $c$ i.e., $1:1$ comparison, match score (e.g., cosine similarity) is computed between the probe and the samples in the gallery $\bm{G}$ corresponding to the identity $c$. The encrypted scores are sent back to the client for decryption and further processing.

\subsection{Encrypted Template Fusion and Matching}
We now describe the various components of template fusion and match score computation. This includes (i) choice of the data encoding scheme, (ii) concatenating two ciphertexts, (iii) efficient ciphertext matrix-plaintext matrix multiplication for linear projection, and (iv) efficient and accurate approximate normalization.

\subsubsection{Input Encoding and Vector Packing\label{sec:encoding}}
\noindent\textbf{Input Encoding:} Before any computation can be performed on encrypted data, an encoding scheme must be selected to enable encryption and arithmetic operations on the resulting ciphertext. The efficiency of ciphertext operations is critically dependent on the encoding scheme chosen to represent features. As such, outline two different encoding schemes for the feature vectors, each of which is better suited for operating either at a small or large scale. \underline{\emph{Dense:}} Encodes each feature vector as a plaintext before encryption, thereby resulting in $n$ ciphertexts. \underline{\emph{SIMD:}} Encodes each dimension of the feature vector as a plaintext before encryption, resulting in $\delta$ ciphertexts.

\vspace{4pt}
\noindent\textbf{Vector Packing:} FHE schemes such as CKKS support arithmetic operations directly on vectors by packing multiple numbers into different slots within a single polynomial. And, in most practical applications, the dimensionality of  feature vectors is much less than the number of available polynomial slots. In such cases, multiple feature vectors can be batched into a single polynomial. The batching allows for SIMD (single instruction multiple data) operations and helps amortize runtime across multiple feature vectors.

Suppose we wish to encode $n$ vectors into polynomials with $m$ slots each. In the dense encoding scheme, $\lceil \frac{n}{\lfloor \frac{m}{\delta} \rfloor} \rceil$ many polynomials are needed if rotation operations are not needed. However, ciphertext template fusion requires rotation operations. So, we pack an extra copy of each vector to simulate the ``wrapping” effect of rotation. Therefore, $\lceil \frac{n}{\lfloor \frac{m}{2\delta} \rfloor} \rceil$ polynomials are needed. In the SIMD encoding scheme, a single dimension of the $n$ vectors can be packed into a single polynomial. In this scheme, $\delta \lceil \frac{n}{m} \rceil$ polynomials are needed to represent $n$ $\delta$-dimensional vectors.

\subsubsection{Concatenating Ciphertexts}
\begin{figure}[h]
    \centering
    \includegraphics[width=0.95\columnwidth]{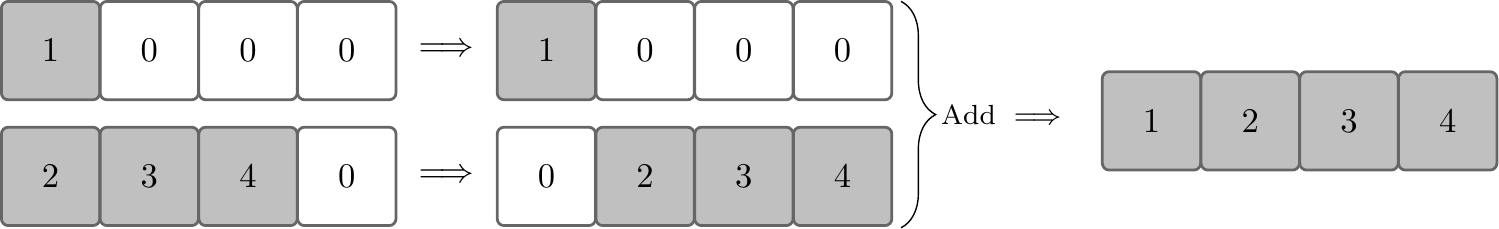}
    \caption{Ciphertext concatenation via rotation and addition for the dense encoding scheme. The second ciphertext (bottom) is right-rotated $\alpha$ slots and added to the first ciphertext (top).\label{fig:concatenation}\vspace{-3pt}}
\end{figure}
The concatenation mechanism depends on our choice of data encoding scheme. \underline{\emph{Dense:}} In this case, each vector in the multibiometric dataset $(\bm{X},\bm{Y})$ is zero-padded before encryption to a dimensionality of $\delta$. Now, concatenation can be done in the encrypted domain by right-rotating each ciphertext in $\bm{Y}$ by $\alpha$ slots and adding to the corresponding ciphertext in $\bm{X}$. \underline{\emph{SIMD:}} As each dimension of the query is packed into a single ciphertext, there is no need to concatenate the features. Instead, simply storing the ciphertexts in a single ordered array is sufficient in this representation.

\subsubsection{Encrypted Linear Projection}
Executing fusion through linear projection requires a matrix-matrix multiplication. Since we learn our projection matrix in the unencrypted domain, the multiplication is a plaintext-ciphertext multiplication, which is considerably more efficient than a ciphertext-ciphertext multiplication. Next, we outline two matrix-vector multiplication techniques, one that is better suited for small-scale datasets and the other for large-scale datasets. However, due to our ciphertext packing scheme, these methods functionally become matrix-matrix algorithms and can be treated as such. Furthermore, we note that it is desirable for the fused representations to be as compact as possible, i.e., $\gamma$ should be small to ease the computational burden of any downstream tasks that are performed directly on the ciphertexts. Hence, the projection matrix $\bm{P}\in\mathcal{R}^{\gamma \times \delta}$ is rectangular.

\begin{figure}[h]
    \centering
    \includegraphics[width=0.95\columnwidth]{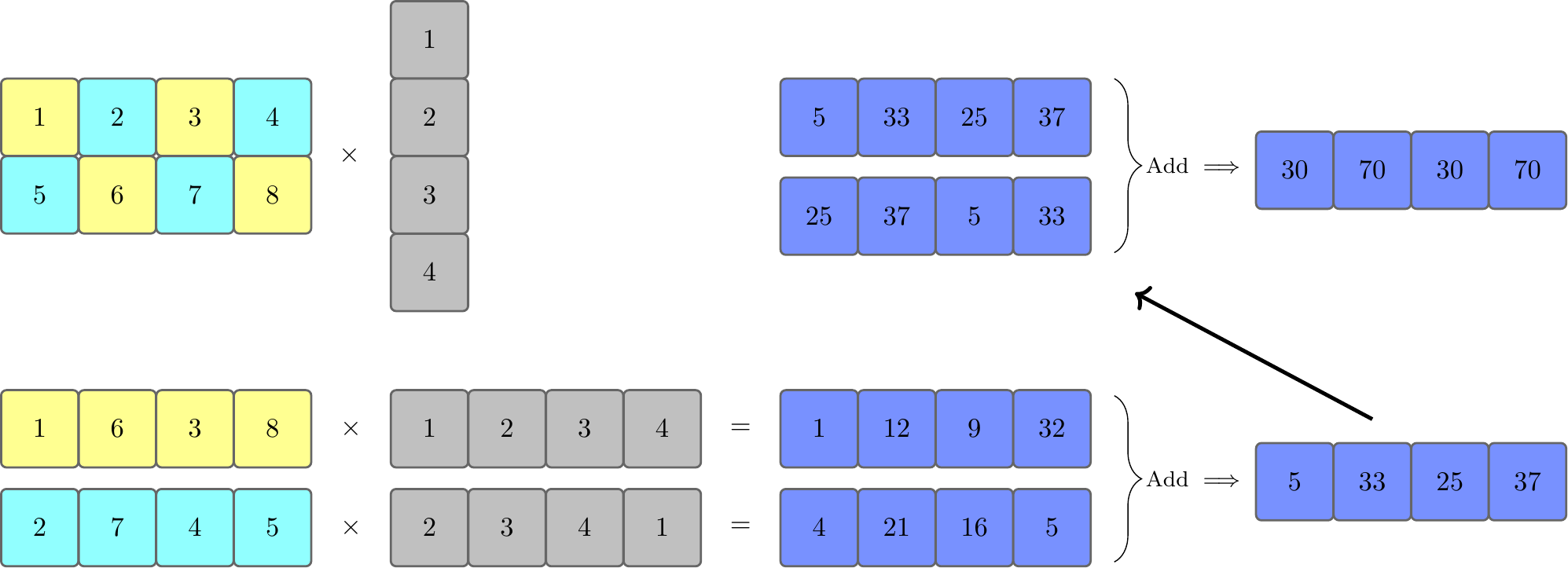}
    \caption{\textbf{Hybrid:} The efficiency of matrix-vector multiplications can be improved through a diagonal encoding scheme for the projection matrix $(\bm{P})$. The query is rotated once and multiplied with each diagonally encoded component of $\bm{P}$. The sum of these results is rotated and added with itself to obtain the final output.\label{fig:hybrid}}
    \vspace{-3pt}
\end{figure}
\noindent\textbf{Hybrid:} When the query vectors are encoded using the dense scheme, the projection matrix can be encoded through a diagonal encoding scheme for efficient matrix-vector products. This scheme, shown in Fig. \ref{fig:hybrid}, was introduced by Juvekar et al. \cite{juvekar2018gazelle} and is specialized for short and wide rectangular matrices, i.e., $\gamma < \delta$. These diagonals are multiplied by rotated versions of the query vector, and the resultant vectors can simply be additively combined to yield the desired matrix-vector multiplication result. This method is best suited for cases where $n$ is small.

\begin{figure}[h]
    \centering
    \includegraphics[width=0.95\columnwidth]{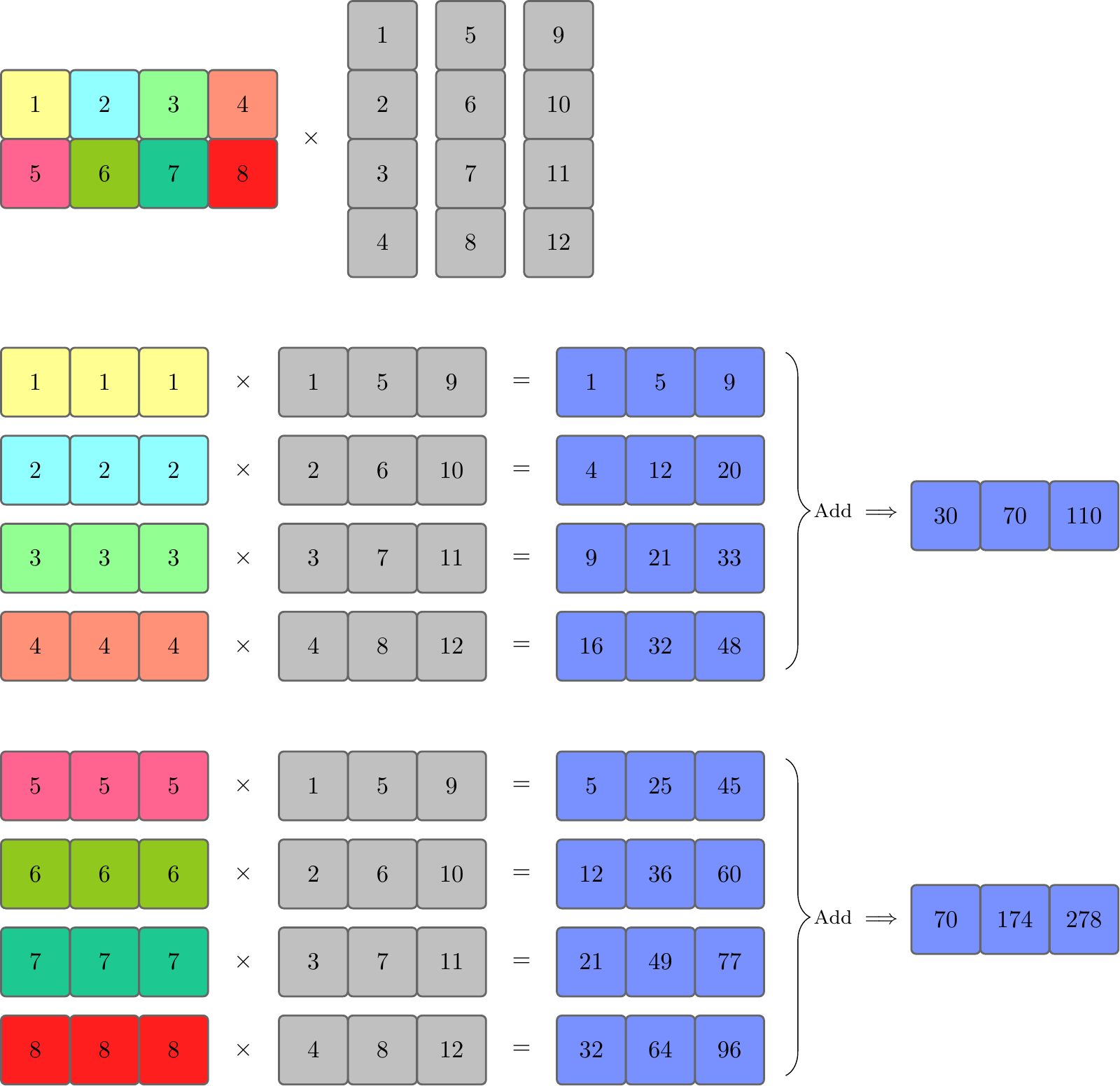}
    \caption{\textbf{SIMD:} This method repeats and encodes each element of the projection matrix as plaintext and multiplies with SIMD encoded query vectors. The result is a single ciphertext for each dimension of the result. This method is best suited for large $n$.\label{fig:simd}}
    \vspace{-3pt}
\end{figure}
\noindent\textbf{SIMD:} When the query vectors are encoded using the SIMD scheme, the projection matrix can also be in a repeated SIMD manner to support scalable matrix-vector products for large $n$. The scheme, shown in Fig.~\ref{fig:simd}, was adopted by Engelmsa~\cite{engelsma2022hers} for scaling search over an encrypted database. This method takes $\gamma \delta$ plaintext-ciphertext multiplications for a single matrix-vector multiplication but admits greater ciphertext packing potential, making it a computationally more efficient solution when $n >> \gamma \delta$. This method also negates the need for any expensive ciphertext rotations. The SIMD scheme, however, is more memory intensive due to the need for loading many plaintexts and ciphertexts in memory as seen in Fig. \ref{fig:MemoryComparison}.

\begin{table*}[t]
\resizebox{2\columnwidth}{!}{%
\centering
\begin{tabular}{l c c c c c c c} 
 \toprule
 \multirow{ 2}{*}{Encoding} & \multirow{ 2}{*}{Operation} & \multicolumn{5}{ c }{Time Complexity} & \multirow{ 2}{*}{Space Complexity} \\
 \cmidrule(lr){3-7}
 
  & & Additions & Plain-Cipher Mult. & Cipher-Cipher Mult. & Mult. Depth & Rotations &  \\ [0.1ex] 
 
 \midrule
 
 \multirow{ 5}{*}{Dense} & Concatenation & $\lceil \frac{n}{l} \rceil$ & 0 & 0 & 0 & $\lceil \frac{n}{l} \rceil$ & $O(p \lceil \frac{n}{l} \rceil)$ \\
 
 & Projection & $(\gamma + log(\delta) - log(\gamma) - 2) \lceil \frac{n}{l} \rceil$ & $\gamma \lceil \frac{n}{l} \rceil$ & 0 & 1 & $(\gamma + log(\delta) - log(\gamma) - 1) \lceil \frac{n}{l} \rceil$ & $O(\gamma p + p \lceil \frac{n}{l} \rceil)$ \\
 
 & Normalization & $log(\gamma) \lceil \frac{n}{l} \rceil$ & $d \lceil \frac{n}{l} \rceil$ & $2 \lceil \frac{n}{l} \rceil$ & $2+d$ & $log(\gamma) \lceil \frac{n}{l} \rceil$ & $O(p \lceil \frac{n}{l} \rceil)$ \\
 
 & Preprocessing & $\lceil \frac{n}{l} \rceil-\lceil \frac{n \gamma}{m} \rceil$ & $\lceil \frac{n}{l} \rceil$ & 0 & 1 & $\lceil \frac{n}{l} \rceil-\lceil \frac{n \gamma}{m} \rceil$ & $O(p \lceil \frac{n}{l} \rceil + p \lceil \frac{n \gamma}{m} \rceil)$ \\
 
 & Matching & $log(\gamma) \lceil \frac{n \gamma}{m} \rceil$ & 0 & $\lceil \frac{n \gamma}{m} \rceil$ & 1 & $log(\gamma) \lceil \frac{n \gamma}{m} \rceil$ & $O(p \lceil \frac{n \gamma}{m} \rceil)$ \\
  
 \midrule
 \multirow{ 5}{*}{SIMD} & Concatenation & - & - & - & - & - & - \\
 
 & Projection & $\gamma (\delta - 1) \lceil \frac{n}{m} \rceil$ & $\delta \gamma \lceil \frac{n}{m} \rceil$ & 0 & 1 & 0 & $O(\delta \gamma p + \gamma p \lceil \frac{n}{m} \rceil)$ \\
  
 & Normalization & $(\gamma-1)\lceil \frac{n}{m} \rceil$ & $d\lceil \frac{n}{m} \rceil$ &$2 \gamma\lceil \frac{n}{m} \rceil$ & $2+d$ & 0 & $O(\gamma p \lceil \frac{n}{m} \rceil)$ \\
 & Preprocessing & - & - & - & - & - & - \\
 & Matching & $(log(\gamma)-1) \lceil \frac{n}{m} \rceil$ & 0 & $ \gamma \lceil \frac{n}{m} \rceil$ & 1 & 0 & $O(\gamma p \lceil \frac{n}{m} \rceil)$ \\
   
 \bottomrule
 
\end{tabular}}
\caption{Time and memory complexity comparison of the Dense and SIMD encoding schemes for template fusion and matching. A preprocessing step is used in the Dense encoding scheme to reduce the number of ciphertexts in the gallery to enable faster matching. $\gamma$ is the output dimensionality of the resultant vector. $\delta$ is the dimensionality of the query vector. For $m$ slots available in a single ciphertext, we define $l=\lfloor \frac{m}{2\delta} \rfloor$. Depending on the encoding scheme, to process $n$ samples, we must perform each operation $\lceil \frac{n}{l} \rceil$ or $\lceil \frac{n}{m} \rceil$ times ($\lceil \frac{n \gamma}{m} \rceil$ times to perform matching in the Dense scheme). $p$ denotes the amount of space a single ciphertext occupies in memory.\label{table:theoretical_complexity}\vspace{-3pt}}
\end{table*}

\subsubsection{Approximate Normalization\label{sec:normalization}}
Recent biometric representations (e.g., DeepPrint~\cite{engelsma2019learning}, ArcFace~\cite{deng2019arcface}) are typically projected to the surface of a unit $\ell_2$-ball\footnote{Other norms like $\ell_1$ or $\ell_{\infty}$ can also be supported by \ourmethod{} if desired.}. Formally, $\hat{u}=\frac{u}{||u||_2}$ where $||u||_2 = \sqrt{\sum_{i=1}^d u_i^2}$ for $u \in \mathbb{R}^d$. This normalization allows for computing cosine similarity simply through a dot-product between a pair of vectors. Such a normalization operation, however, cannot be performed directly on the ciphertexts since FHE schemes do not support non-arithmetic operations such as square root and division in the encrypted domain. Although it is possible to approximate each of these operations individually~\cite{cheon2019numerical, babenko2021euclidean}, the computational efficiency can be significantly improved by directly approximating the inverse square root operation. Panda \cite{panda2021principal} showed it is possible to approximate inverse square root through the iterative Goldschmidt's Algorithm \cite{cetin2015arithmetic, markstein2004software} but is impractical for our purposes due to its high multiplicative depth.

We adopt a composite polynomial of the form $f(x) = (g_k \circ g_{k-1} \circ \cdots \circ g_1)(x)$, where each $g_i(x)$ is a low-degree polynomial, to approximate the inverse square root function in a desired interval of $x$ i.e., $\frac{1}{\sqrt{x}} \approx f(x) \forall x \in [a,b]\footnote{See the appendix for discussion on choosing the interval.}$. The number of composite functions $k$ and the degree of each $g_i$ determine the homomorphic multiplicative depth of the operation. Higher degree polynomials offer a better approximation of this function, but also increase the multiplicative depth of the circuit. Hence, there is a trade-off between accuracy of the approximation and computational efficiency.

\begin{figure*}[h]
    \centering
    \begin{subfigure}{0.24\textwidth}
        \centering
        \includegraphics[width=\textwidth]{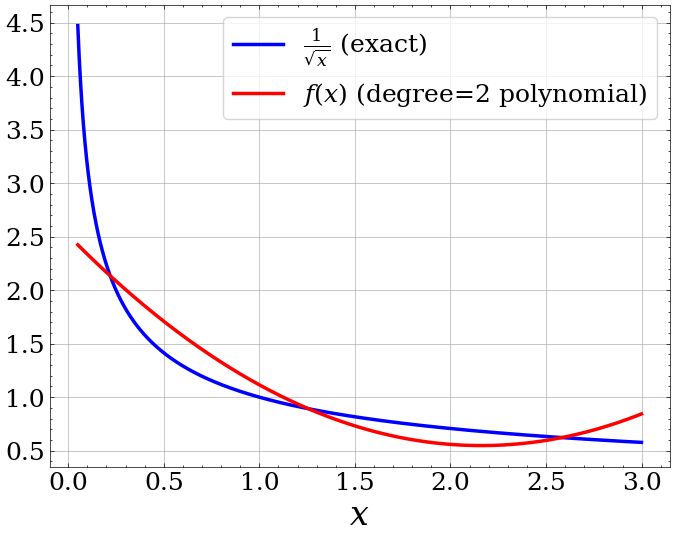}
    \end{subfigure}
    \begin{subfigure}{0.24\textwidth}
        \centering
        \includegraphics[width=\textwidth]{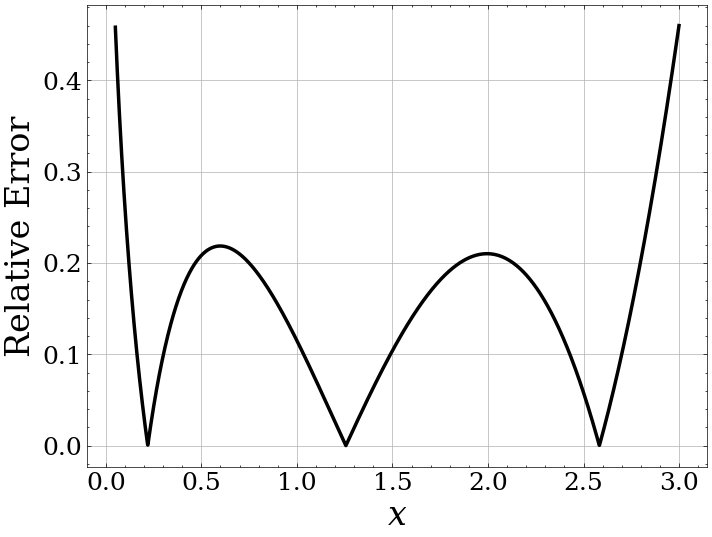}
    \end{subfigure}
    \begin{subfigure}{0.24\textwidth}
        \centering
        \includegraphics[width=\textwidth]{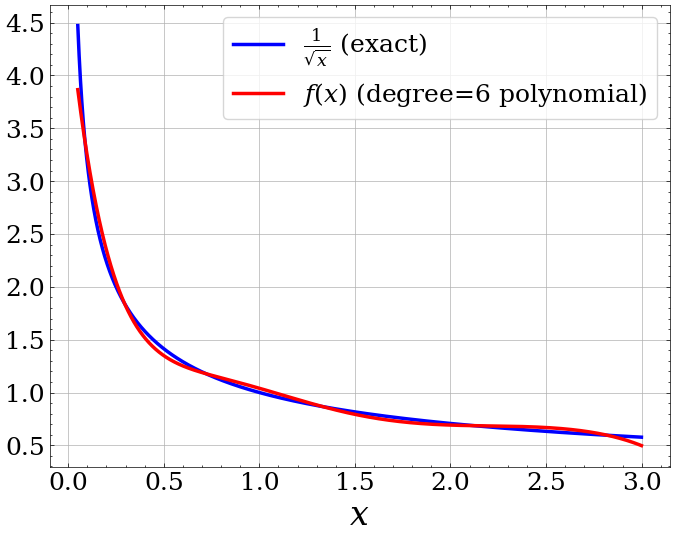}
    \end{subfigure}
    \begin{subfigure}{0.24\textwidth}
        \centering
        \includegraphics[width=\textwidth]{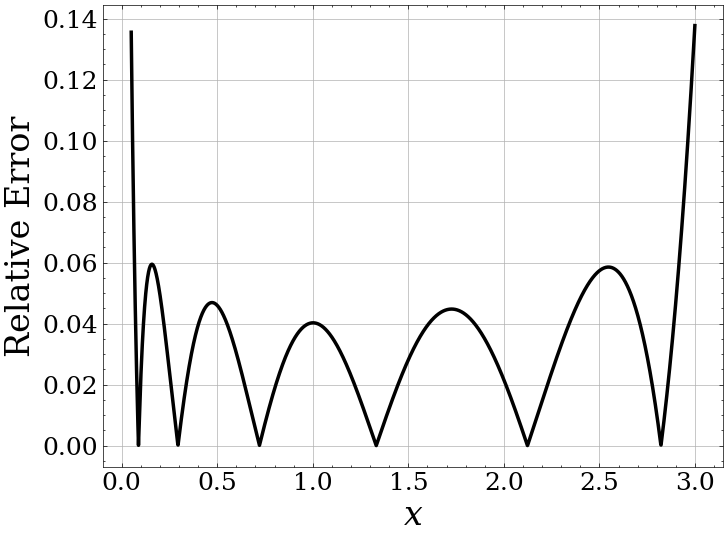}
    \end{subfigure}
    \caption{Polynomial approximations of inverse square root over the interval [0.05, 3.0] for polynomials of degree 2 and 6. Relative error $\left(\left|f(x)-\frac{1}{\sqrt{x}}\right|/\left|\frac{1}{\sqrt{x}}\right|\right)$ of the approximations are shown to the right of each plot.\label{fig:PolynomialDegree3}\vspace{-3pt}}
\end{figure*}

\subsubsection{Computational Complexity}
Table \ref{table:theoretical_complexity} shows an analytical comparison of the time and space complexity of the end-to-end pipeline for both the Dense and SIMD encoding schemes. We show the required number of atomic operations for each stage of the pipeline including concatenation, projection, normalization, preprocessing (that is necessary for matching), and matching.

\subsection{FHE Aware Learning of Projection Matrix}
Having described the inference process, we now turn our attention to learning the optimal linear projection matrix $\bm{P}$ for template fusion. We posit that $\bm{P}$ can be learned in the unencrypted domain using biometric templates that are already available and hence do not suffer from privacy concerns. And, once learned, it can be employed for fusing the templates from private data for inference.

The projection matrix should map vectors of the same class close together for a given distance metric, while those of different classes should be far apart. To realize this goal, we adapt the concept of the maximum-margin loss function introduced by Siena et al. \cite{siena2013maximum} for learning $\bm{P}$. The loss function minimizes the distance between samples of the same class and uses a hinge loss on triplets of samples involving a similar and dissimilar pair. We build upon this concept and adapt it in several ways to satisfy the unique combination of constraints imposed by the multi-modal fusion of features from deep neural networks and those of normalization approximations induced by FHE computations at inference. 

Firstly, we adapt the loss function for multimodal feature-level fusion. Specifically, unlike Siena et al.~\cite{siena2013maximum} who seek to learn a pair of projection matrices with Euclidean distance-based metric, we learn a single projection matrix with cosine similarity\footnote{Note that \ourmethod{} can also optimize for Euclidean distance if desired.} based metric. Given a concatenated dataset $\tilde{\bm{X}}$, the loss function is defined as:
\begin{equation}
\label{eq:loss}
\begin{aligned}
\mathcal{L} &= \lambda \frac{\sum_{M} d(\tilde{\bm{x}}_i,\tilde{\bm{x}}_j)}{|M|} + \\
& (1-\lambda) \frac{\sum_{V} [m+d(\tilde{\bm{x}}_i,\tilde{\bm{x}}_j)-d(\tilde{\bm{x}}_i,\tilde{\bm{x}}_k)]_+}{|V|}
\end{aligned}
\end{equation}
\noindent where $d(\tilde{\bm{x}}_i,\tilde{\bm{x}}_j) =  1 - \frac{(\bm{P}\tilde{\bm{x}}_i)^T(\bm{P}\tilde{\bm{x}}_j)}{\|\bm{P}\tilde{\bm{x}}_i\|\|\bm{P} \tilde{\bm{x}}_j\|}$, $[x]_+ = max(0,x)$, $M$ is the set of all pairs in $\tilde{\bm{X}}$ belonging to the same class, $V$ denotes the set of all triplets $(\bm{x}_i,\bm{x}_j,\bm{x}_k)$ such that $(\bm{x}_i,\bm{x}_j)$ belong to the same class and $(\bm{x}_i,\bm{x}_k)$ belong to different classes, $\lambda$ is a hyperparameter that weighs the ``push" and ``pull" terms' influences on the loss, $m$ is the margin hyperparameter that determines the desired margin of separation between samples belonging to the same class and those belonging to different classes. The margin hyperparameter used in the triplet hinge loss can appropriately take on any value in the range $\left[0,\frac{c}{c-1}\right]$ for $c$ classes \cite{wang2018cosface}.

Secondly, we note that the loss function in \eqref{eq:loss} is defined with exact normalization, while at the inference stage \ourmethod{} can only perform approximate normalization as described in Sec.~\ref{sec:normalization}. For instance, Fig.~\ref{fig:PolynomialDegree3} shows a comparison between the exact inverse square root function and polynomial approximations of degrees 2 and 6. The mismatch between the unencrypted training and encrypted inference objectives observed here leads to performance degradation, as we demonstrate in Section~\ref{sec:experiments}. To mitigate this loss and recover the matching performance in the unencrypted domain, we incorporate the approximate normalization into the distance metric as,
\begin{equation}
\label{eq:approxdistance}
d(\tilde{\bm{x}}_i,\tilde{\bm{x}}_j) =  1 - (\bm{P}\tilde{\bm{x}}_i \odot f(\bm{P}\tilde{\bm{x}}_i))^T (\bm{P}\tilde{\bm{x}}_j \odot f(\bm{P}\tilde{\bm{x}}_j)) 
\end{equation}
\noindent where $\odot$ is the Hadamard product, and $f(\cdot)$ is the composite polynomial defined in Sec~\ref{sec:normalization} and can be computed efficiently in the encrypted domain. Substituting the distance metric in \eqref{eq:loss} by \eqref{eq:approxdistance} allows the learned projection matrix to compensate for the approximate normalization to a large extent, if not fully eliminate it.


\section{Experiments\label{sec:experiments}}
We evaluate the effectiveness of \ourmethod{} and analyze the effect of our design choices, both in terms of matching accuracy and computational complexity.

\vspace{3pt}
\noindent\textbf{Implementation Details:} To learn the projection matrix, we use the Adam~\cite{kingma2014adam} optimizer with a learning rate of $5 \times 10^{-3}$, a weight decay of $1 \times 10^{-4}$ and train for 1000 epochs. Our encrypted inference is based on the CKKS scheme implemented in Microsoft's SEAL~\cite{sealcrypto} library. Depending on the multiplicative depth of our approximate normalization method, we either use a polynomial modulus degree ($N$) of 16,384 or 32,768 along with a chain of very large prime numbers totaling 420, 580 or 860 bits as the coefficient modulus ($q$).

\begin{figure*}[h]
    \centering
    \begin{subfigure}{0.245\textwidth}
        \centering
        \includegraphics[width=\columnwidth]{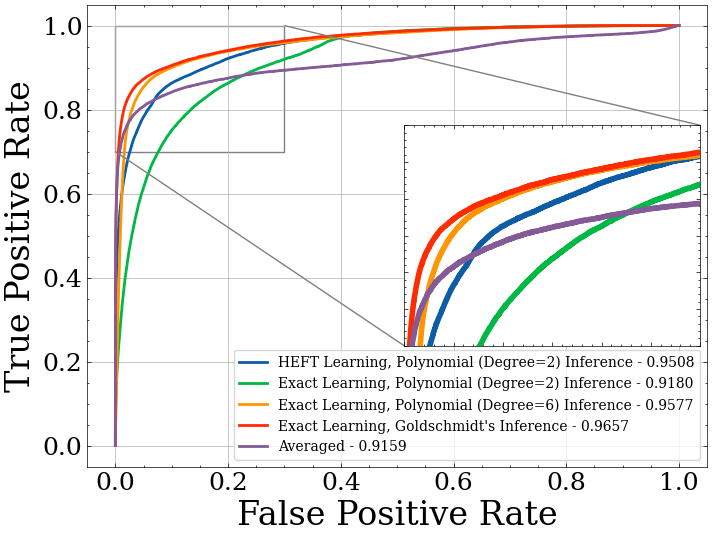}
        \caption{\label{fig:ROC}}
    \end{subfigure}
    \begin{subfigure}{0.245\textwidth}
        \centering
        \includegraphics[width=\columnwidth]{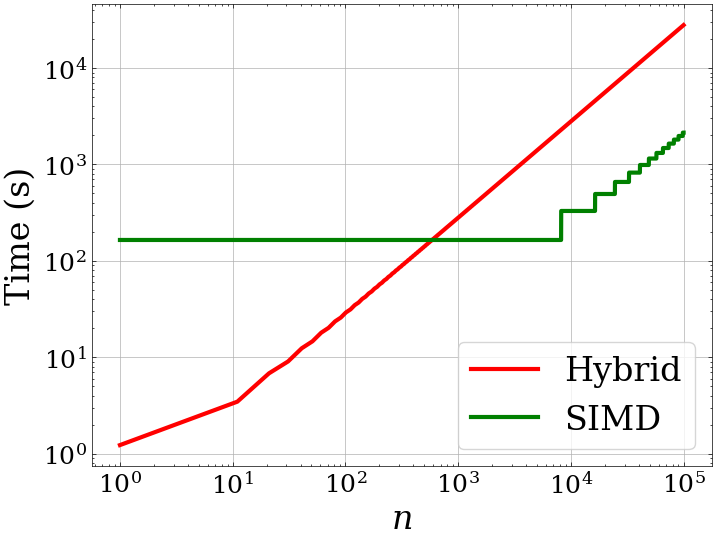}
        \caption{\label{fig:TimeComparison}}
    \end{subfigure}
    \begin{subfigure}{0.245\textwidth}
        \centering
        \includegraphics[width=\columnwidth]{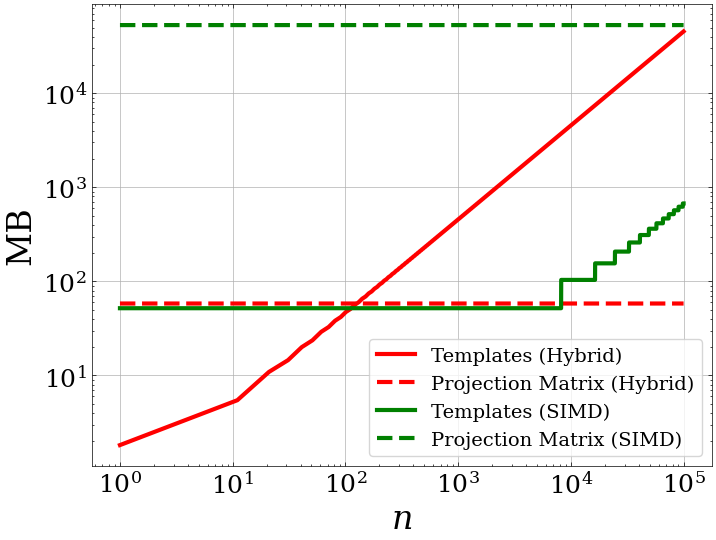}
        \caption{\label{fig:MemoryComparison}}
    \end{subfigure}
    \begin{subfigure}{0.245\textwidth}
        \centering
        \includegraphics[width=\columnwidth]{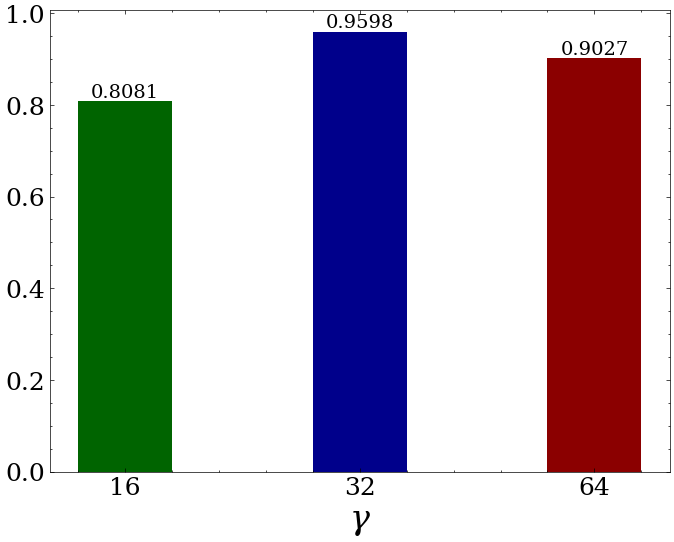}
        \caption{\label{fig:ablation_gamma}}
    \end{subfigure}
    \caption{(a) ROC comparison of \ourmethod{} against baselines. (b) and (c) Comparison of theoretical runtimes and memory requirements for Hybrid and SIMD encoding schemes with $\delta=1024$ and $\gamma=32$. (d) Ablation study on $\gamma$, where 32 performs the best in our case.}
\end{figure*}

\subsection{Evaluation Datasets}

\noindent\textbf{\href{https://ai.googleblog.com/2017/08/launching-speech-commands-dataset.html}{Google Speech Commands}:} This dataset comprises spoken single-word commands from many speakers. We use 5380 samples over 188 classes. We extract 512-dimensional feature vectors with the Deep Speaker~\cite{li2017deep} model, which is trained on the train-clean-360 portion of the LibriSpeech \cite{panayotov2015librispeech} dataset using a publicly available \href{https://github.com/philipperemy/deep-speaker}{implementation}.

\vspace{3pt}
\noindent\textbf{CPLFW \cite{CPLFWTech}:} This benchmark face dataset is a harder version of LFW that incorporates cross-posed faces. We extract 512-dimensional feature vectors from a pre-trained \href{https://github.com/rcmalli/keras-vggface}{VGG16 model trained on VGGFace}\cite{simonyan2014very, parkhi2015deep}.

We pair 2 samples of 188 identities from CPLFW with those in the Google Speech Commands Dataset to create a multimodal dataset. This results in 10,760 samples over 188 classes as our dataset. Of these, 20\% of the classes are used for testing, 20\% for validation, and 60\% for training. This yields a test set of 1028 samples.

\subsection{Comparison and Selection of Encoding Scheme}
As discussed in Sec.~\ref{sec:encoding} there are two encoding schemes, each with different computational properties. To select the one that is appropriate for our purposes, we first numerically compare them. The time and space complexity for the end-to-end pipeline, i.e., concatenation, projection, approximate normalization, and match score computation, of each encoding scheme are shown in Figs.~\ref{fig:TimeComparison} and~\ref{fig:MemoryComparison} respectively. To compute the numerical values from the theoretical expressions in Table \ref{table:theoretical_complexity}, we compute the runtime of each atomic operation in SEAL by averaging over 1,000 operations with the appropriate encryption parameters. Similarly, space is calculated by examining the size of a single ciphertext. As expected, we observe a cross-over point between the two, with SIMD being more efficient in terms of latency for $n > 1000$ and in terms of memory for $n > 10000$. Furthermore, for our dataset of size 1028, while the latency between the two is comparable, the dense encoding scheme has lower memory requirements. Therefore, we use the dense encoding scheme for all further experiments.

\subsection{Evaluation Metrics and Results}
In \ourmethod{} to compute the cosine similarity of feature vectors, we apply the appropriate normalization method (exact or approximate) on each vector and then take their dot product. Finally, we use the AUROC metric to evaluate the template fusion methods. The metric is computed in the unencrypted domain after decrypting the match scores.

\begin{table}[t]
\resizebox{\columnwidth}{!}{%
\centering
\begin{tabular}{c c c c c c c} 
\toprule
\multirow{ 2}{*}{Index} & \multirow{ 2}{*}{Data} & \multirow{ 2}{*}{Domain} & \multicolumn{2}{c}{Normalization} & \multirow{ 2}{*}{Dimensionality} & \multirow{ 2}{*}{AUROC} \\ [0.1ex] 
 \cmidrule(lr){4-5}
 & &  & Inference & Learning & &  \\ [0.1ex] 
\midrule
1 & CPLFW & Unencrypted & Exact & - & 512 & 0.8401 \\
2 & GSC & Unencrypted & Exact & - & 512 & 0.8550 \\
3 & Average~\cite{drozdowski2021feature} & Encrypted & Exact & - & 512 & 0.9159 \\
4 & Concatenation & Unencrypted & Exact & - & 1024 & 0.9253 \\
5 & Learned & Unencrypted & Exact & Exact & 32 & 0.9755 \\
6 & Learned & Encrypted & Poly (Deg=6) & Exact & 32 & 0.9577 \\
7 & Learned & Encrypted & Poly (Deg=2) & Exact & 32 & 0.9180 \\
8 & Learned & Encrypted & Goldschmidt's & Exact & 32 & 0.9657 \\
9 & Learned & Unencrypted & Exact & \ourmethod{} (Deg=2) & 32 & 0.9598 \\
10 & Learned & Encrypted & Poly (Deg=2) & \ourmethod{} (Deg=2) & 32 & 0.9508 \\
\bottomrule
\end{tabular}}
\caption{AUROC comparison of \ourmethod{} versus baselines\label{table:AUC_results}\vspace{-3pt}}
\end{table}

\vspace{3pt}
\noindent\textbf{Matching Performance:} To evaluate the performance of \ourmethod{} we compare it against the following baselines, i) the unibiometric templates, ii) a simple concatenation of the unibiometric features, i.e., $\tilde{\bm{X}}$, iii) training using exact normalization, and iv) the feature averaging fusion technique introduced in \cite{drozdowski2021feature}. Table \ref{table:AUC_results} compares the performance of \ourmethod{} with the baselines. We make the following observations, i) all fusion techniques with exact normalization (rows 1-5) namely averaging, concatenation and learned projection improve biometric matching performance with the latter providing the best performance, ii) approximate normalization at inference leads to drop in performance (rows 5 \& 6, 5 \& 7) ii) higher degree polynomial for approximate normalization performs better than the lower degree counterpart (row 6 \& 7), iv) learning the projection matrix by taking the approximate normalization into account helps recover performance (rows 7 \& 10), v) approximate normalization through Goldschmidt's algorithm is more accurate than that using polynomials (rows 6 \& 8, 7\& 8) and vi) computing the match score in the encrypted domain entails a slight loss in performance (rows 9 \& 10). Overall, \ourmethod{} improves AUROC by 11.07\% and 9.58\% over CPLFW and Google Speech Commands, respectively.

\vspace{3pt}
\noindent\textbf{Computational Complexity:} The efficiency of homomorphic operations is critically dependent on the chosen encryption parameters. We select these parameters based on the multiplicative depth needed for end-to-end fusion and matching. Table \ref{table:time_results2} shows the latency of each individual component of \ourmethod{}. First, we observe a trade-off between performance and time complexity, with the $2^{nd}$-degree polynomial being $2\times$ faster than the $6^{th}$-degree polynomial for enrollment. Although Goldschmidt's algorithm performs the best, it is $4.8\times$ and $9.6\times$ slower than \ourmethod{} with degree two approximation for enrollment and authentication respectively.

\vspace{3pt}
\noindent\textbf{Ablation Study:} We study the effect of $\gamma$, the dimensionality of the fused templates. Noting that $\gamma$ should be a power of two to enable efficient match score computation, we compare three choices. Figure~\ref{fig:ablation_gamma} shows that $\gamma=32$ yields the best performance, with 64 being slightly better than 16.

\begin{table}[t]
\resizebox{\columnwidth}{!}{%
\centering
\begin{tabular}{l c c c c c c c} 
\toprule
Protocol & Enc. Norm. Method & Concatenation & Projection & Normalization & Preprocessing & Fusion Total & Score Comp. \\ [0.1ex] 
 \midrule

\multirow{3}{*}{Enrollment} & Poly (Deg=2) & 5.68 & 244.89 & 31.40 & 3.41 & 285.38 & -\\
& Poly (Deg=6) & 11.17 & 470.86 & 83.32 & 3.62 & 568.97 & -\\
& Goldschmidt's & 23.22 & 954.03 & 380.28 & 2.31 & 1,359.84 & -\\

\midrule
\multirow{3}{*}{Authentication} & Poly (Deg=2) & 22.72 & 979.54 & 125.59 & - & 1,127.85 & 4.87\\
& Poly (Deg=6) & 89.05 & 3,752.24 & 663.95 & - & 4,505.24 & 5.21\\
& Goldschmidt's & 185.00 & 7,602.64 & 3,030.47 & - & 10,818.11 & 2.75\\

\bottomrule
\end{tabular}}
\caption{Time (milliseconds) breakdown for each step in enrollment and authentication  for a single sample. For comparison the same operations in message-space takes 0.62, 1.02, 11.75, and 4.51 $\mu$s respectively for concatenation, projection, normalization, and score computation per sample/match. \label{table:time_results2}}
\end{table}

\section{Conclusions\label{sec:conclusions}}
In this paper, we proposed \ourmethod{}, the first non-interactive end-to-end homomorphically encrypted multimodal feature-level fusion and matching system. From an inference perspective, we carefully analyzed different data encoding and linear projection schemes and introduced approximate scale normalization through composite polynomial. From a learning perspective, we introduced FHE-Aware learning that explicitly accounts for the inherent limitations of FHE, namely the inability to perform exact normalization. Experimental results show that \ourmethod{} can overcome the performance losses due to approximations induced by FHE constraints and improve performance over the unibiometric features by 11.07\% and 9.58\% AUROC respectively while being practically feasible, taking 884 ms for fusing a pair of 512-dimensional vectors and matching against a gallery of 1024 templates.

\vspace{3pt}
\noindent\textbf{Acknowledgments:} This material is based upon work supported by the Center for Identification Technology Research and the National Science Foundation under Grant No. 1841517.
\begin{appendices}

This appendix includes the following:
\begin{enumerate}
    \item \textbf{Section~\ref{sec:additionallinear}:} Two additional linear projection methods.
    \item \textbf{Section~\ref{sec:practical}:} Discussion on practical considerations to account for when approximating inverse square root in the encrypted domain.
    \item \textbf{Section~\ref{sec:additionalexperiments}:} Additional experiments on fusing fingerprint and face templates.
\end{enumerate}

\section{Additional Linear Projection Methods\label{sec:additionallinear}}
\vspace{3pt}
\noindent\textbf{Naive:} The naive method~\cite{halevi2014algorithms} aims to perform a matrix-vector multiplication similarly to how it would be computed in the unencrypted domain. The encoding scheme for the matrix encodes each row of the matrix as a plaintext. The query vector is presumed to be in the dense representation. The homomorphic inner product is taken between the vector and each row of the matrix. This results in $\gamma$ ciphertexts. Each of these ciphertexts holds one dimension of the result replicated in every slot. To combine these results into a single ciphertext, each result ciphertext is multiplied by a mask plaintext containing one in the appropriate index and zeroes elsewhere. All of these ciphertexts are added together, yielding the final result. This method is considerably slowed down by the fact that it features two serial multiplications. This necessitates a slower ciphertext-ciphertext multiplication as well as requiring a larger coefficient modulus to be selected, which slows all our operations.
\begin{figure}[h]
    \centering
    \includegraphics[width=0.95\columnwidth]{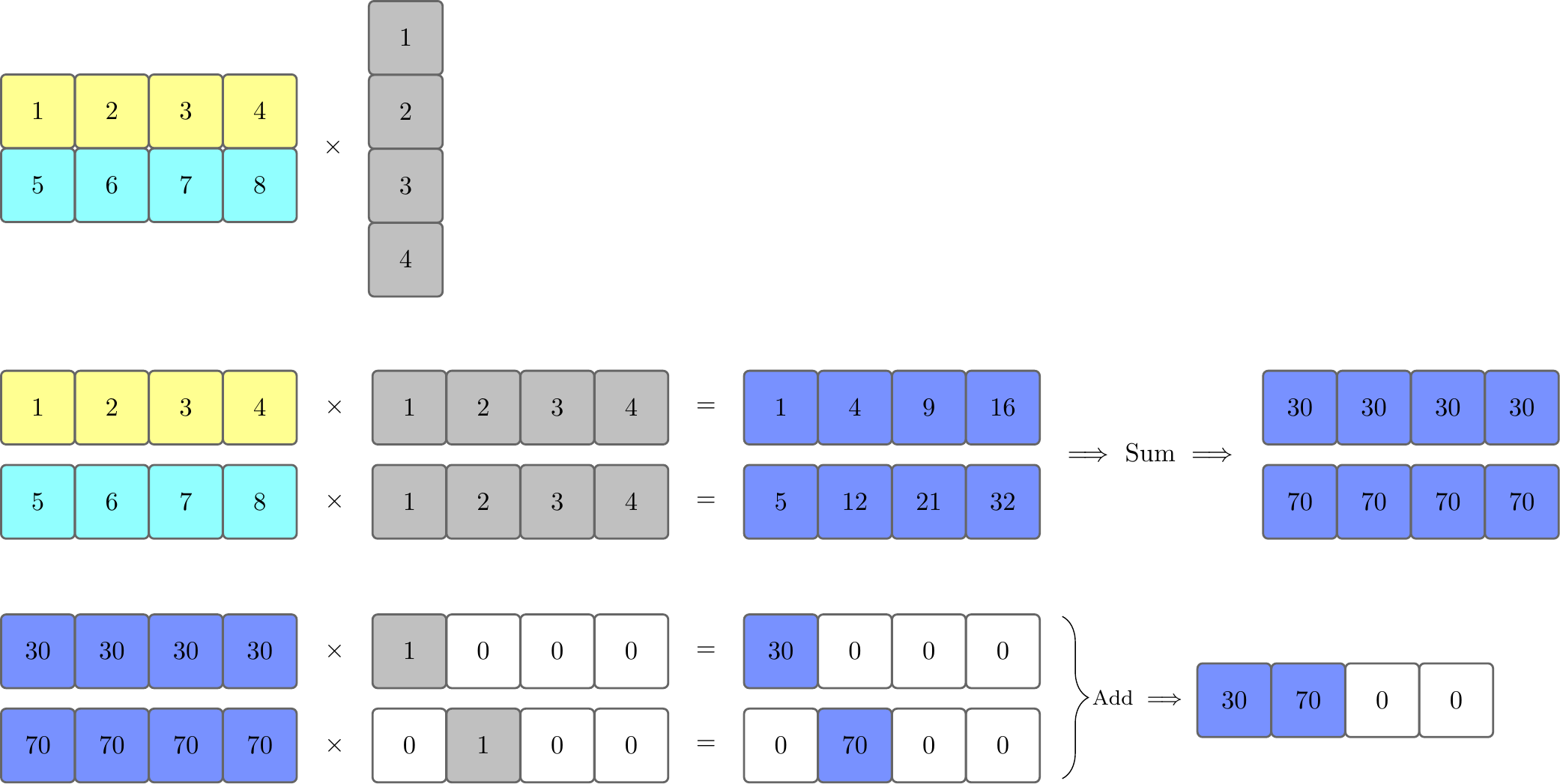}
    \caption{\textbf{Naive method:} A naive method for matrix-vector multiplication. Each row of the projection matrix is encoded as a plaintext. Through a series of homomorphic inner products, we obtain a ciphertext representing each dimension of the result. Through masking and adding, we arrive at the final result.\label{fig:naive}}
\end{figure}

\begin{figure}[h]
    \centering
    \includegraphics[width=0.95\columnwidth]{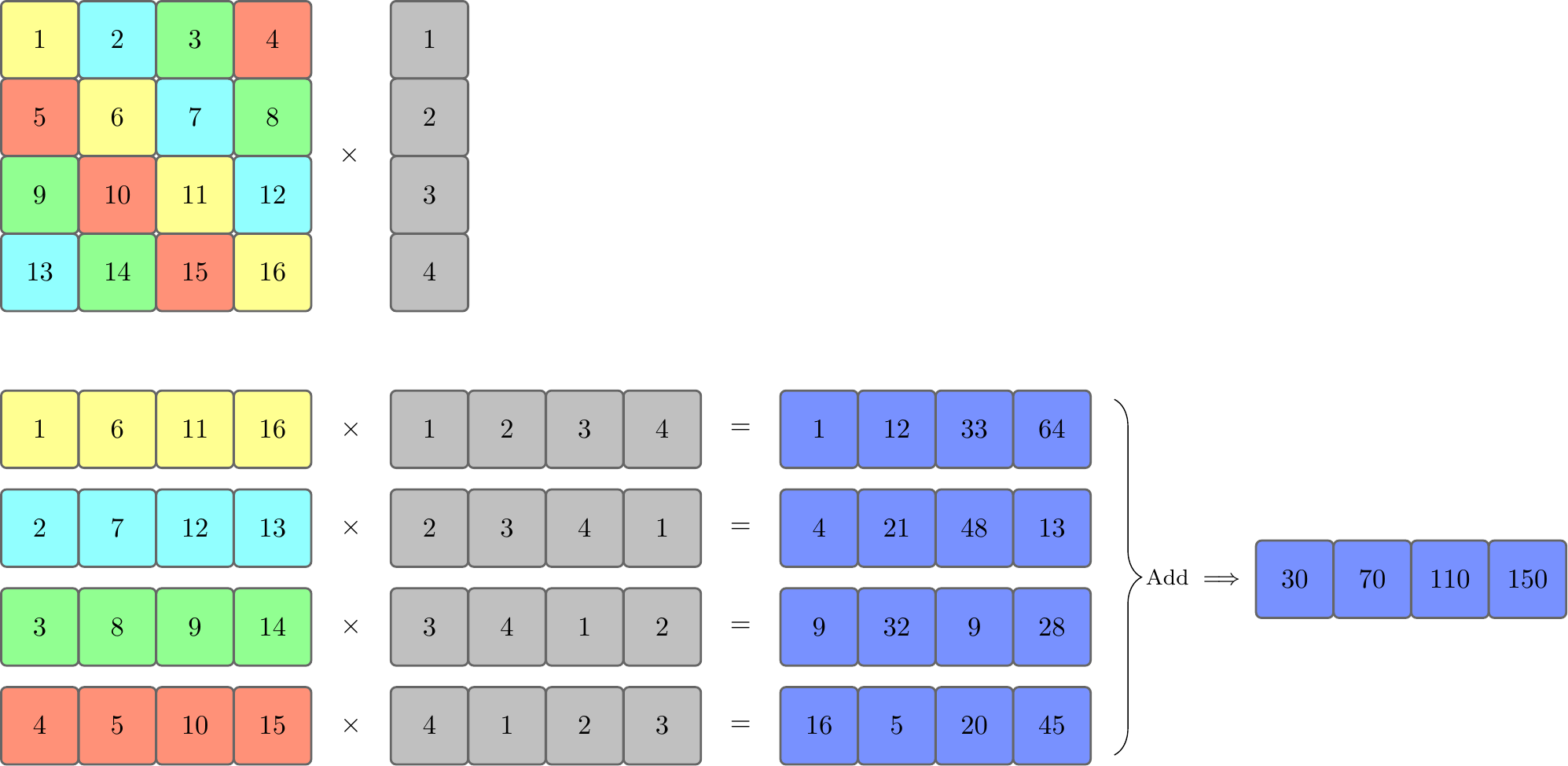}
    \caption{\textbf{Diagonal method:} Encoding over diagonals of a square matrix instead of over the rows results in a method that reduces the multiplicative depth from two to one. The query vector is rotated once and multiplied by each subsequent matrix plaintext. The sum of all these results yields the final result.\label{fig:diagonal}}
\end{figure}

\vspace{3pt}
\noindent\textbf{Diagonal:} This method, devised by Halevi and Shoup~\cite{halevi2014algorithms}, reduces the number of serial multiplications from two to one by not computing homomorphic inner products. Instead, by encoding the diagonals of a matrix as seen in Figure \ref{fig:diagonal} and multiplying by rotated versions of the query vector (which is again presumed to be in the dense encoding scheme), the resultant vectors can simply be summed to yield the result. This diagonal encoding scheme and algorithm assumes that the matrix is square. Thus, we either need to zero-pad our rectangular matrix to use this method or not reduce the dimensionality of the concatenated representation (i.e., $\gamma = \delta$). Compressing the representation is important to enable efficient match score computation, however, so this approach is not desirable.

\begin{figure}[h]
    \centering
    \includegraphics[width=0.95\columnwidth]{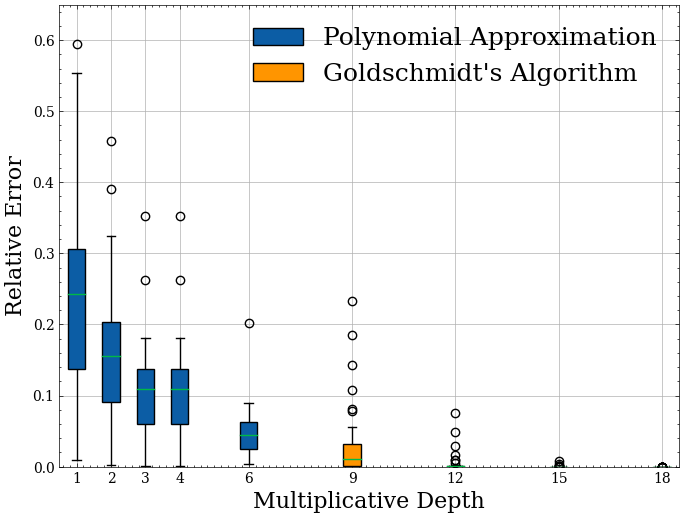}
    \caption{Comparison of multiplicative depth vs. relative error for different selections of inverse square root approximation. Polynomial approaches can achieve lower multiplicative depth, but incur more relative error than Goldschmidt's Algorithm.\label{fig:ErrorComparison}}
\end{figure}

\section{Practical Considerations for Approximating Inverse Square Root\label{sec:practical}}
Panda \cite{panda2021principal} showed that it is possible to normalize an encrypted vector using Goldschmidt's Algorithm to approximate inverse square root. This method of approximating inverse square root in the encrypted domain is not particularly suitable for practical applications due to its high multiplicative depth requirements, slowing every computation performed in the encrypted domain. We instead use a polynomial approximation constrained over an interval. Figure \ref{fig:ErrorComparison} shows the multiplicative depth of different parameters for polynomial and Goldschmidt's approximations and their relative errors. Care must be taken to ensure that an appropriate range is selected for the approximation, such that the results of the projection yield squared norms that fall in the valid range. Smaller ranges yield more precise polynomial approximations, but are more susceptible to samples not falling within the range. We select the range $[0.05-3.00]$ and constrain the norm of our projection matrix $\bm{P}$ such that the mean of the squared norms of the projections within the train set is in the middle of the range. If samples fall outside this valid range during training, we skip those samples during that epoch.

\section{Additional Experiments\label{sec:additionalexperiments}}
We explore the capacity of \ourmethod{} when used on a dataset with little room for performance improvement. NIST SD4 \cite{watson1992nist} is a dataset consisting of ink-rolled fingerprint images. It contains 4,000 samples over 2,000 identities. 192D features are extracted via DeepPrint \cite{engelsma2019learning}. We pair the well-performing NIST SD4 dataset with the more challenging CPLFW dataset to create a multimodal dataset of 7,696 samples over 1,924 classes.

\begin{figure}[h]
    \centering
    \includegraphics[width=\columnwidth]{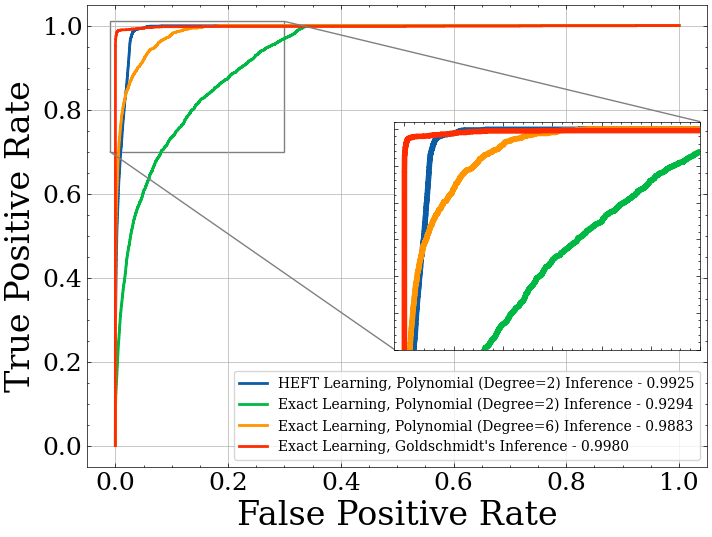}
    \caption{ROC comparison of HEFT against baseline.\label{fig:ROC2}}
\end{figure}

The AUROC of \ourmethod{} compared to the baseline is shown in Table \ref{table:AUC_results2} and Fig. \ref{fig:ROC2}. The averaging method is not evaluated, since it only works on representations that are of the same dimensionality. NIST SD4 is a somewhat compact representation of 192D, but \ourmethod{} can compress it by a factor of 6 to 32D with only a 0.75\% drop in AUROC. Notably, \ourmethod{} improves matching performance over exact learning by 6.31\%. Additionally, \ourmethod{} outperforms the exact normalization learning method when a degree 6 polynomial is used at inference time by 0.42\%. These results show \ourmethod{}'s utility at compressing highly accurate representations via fusion.

\begin{table}[t]
\resizebox{\columnwidth}{!}{%
\centering
\begin{tabular}{c c c c c c} 
\toprule
Data & Domain & Encrypted Normalization & Learning Normalization & Dimensionality & AUROC \\ [0.1ex] 
\midrule
 
CPLFW & Message & - & - & 512 & 0.8253 \\

NIST SD4 & Message & - & - & 192 & 0.99997 \\

Concatenation & Message & - & - & 704 & 0.9982 \\

Learned & Message & - & Exact & 32 & 0.99990 \\

Learned & Encrypted & Poly (Deg=6) & Exact & 32 & 0.9883 \\

Learned & Encrypted & Poly (Deg=2) & Exact & 32 & 0.9294 \\

Learned & Encrypted & Goldschmidt's & Exact & 32 & 0.9980 \\

Learned & Message & - & \ourmethod{} (Deg=2) & 32 & 0.9925 \\

Learned & Encrypted & Poly (Deg=2) & \ourmethod{} (Deg=2) & 32 & 0.9925 \\
\bottomrule
\end{tabular}}
\caption{AUROC comparison of \ourmethod{} versus baseline\label{table:AUC_results2}}
\end{table}

\end{appendices}
{\small
\bibliographystyle{ieee}
\bibliography{egbib}
}
\end{document}